\documentclass{article}
\usepackage{spconf,amsmath,epsfig}

\usepackage{bm}

\usepackage{fancyhdr}
\begin{document}\sloppy
\def\x{{\mathbf x}}
\def\L{{\cal L}}

\title{Multimodal Semantic Attention Network for Video Captioning}

\name{Liang Sun\textsuperscript{1}, Bing Li\textsuperscript{2,3},  Chunfeng Yuan\textsuperscript{2,3}, Zhengjun Zha\textsuperscript{1}, Weiming Hu\textsuperscript{2,3}\thanks{
This work is partly supported by the National Key R\&D Plan (Nos. 2017YFB1002801, 2017YFB1300201 and 2016QY01W0106), the Natural Science Foundation of China (Nos.U1803119, U1736106, 61751212, 61721004, 61772225, 61622211 and 61620106009) as well as the Fundamental Research Funds for the Central Universities under Grant WK2100100030, the NSFC-General Technology Collaborative Fund for Basic Research (Grant No. U1636218), the Key Research Program of Frontier Sciences, CAS (Grant No. YZDJ-SSW-JSC040), Beijing Natural Science Foundation (Nos. JQ18018, L172051 and L182058) and the CAS External Cooperation Key Project. Bing Li is also supported by Youth Innovation Promotion Association, CAS.}}
\address{{\textsuperscript{1}School of Information Science and Technology, University of Science and Technology of China}\\
	{\textsuperscript{2}National Laboratory of Pattern Recognition, Institute of Automation, Chinese Academy of Sciences} \\	
	{\textsuperscript{3}CAS Center for Excellence in Brain Science and Intelligence Technology} \\
	slucius@mail.ustc.edu.cn, zhazj@ustc.edu.cn, \{bli, cfyuan, wmhu\}@nlpr.ia.ac.cn}

\maketitle

\begin{abstract}
Inspired by the fact that different modalities in videos carry complementary information, 
we propose a \textit{Multimodal Semantic Attention Network} (MSAN), which is a new encoder-decoder framework incorporating multimodal semantic attributes for video captioning. In the  
encoding phase, we detect and generate multimodal semantic attributes by formulating it as a multi-label classification problem. Moreover, we add auxiliary classification loss to our model that can obtain more effective visual features and high-level multimodal semantic attribute distributions for sufficient video encoding. 
In the decoding phase, we extend  each 
weight matrix of the conventional LSTM to an ensemble of attribute-dependent weight matrices,  
and employ attention mechanism to pay attention to different attributes at each time of the captioning process. 
We evaluate algorithm on two popular public benchmarks: MSVD and MSR-VTT, achieving competitive results with current state-of-the-art across six evaluation  metrics.
\end{abstract}
\vspace{-0.2in}
\section{Introduction}
\vspace{-0.1in}
\label{sec:intro}

Video captioning refers to the automatic generation of a natural language description that summarizes an input video, it has widespread applications including human-computer interaction, video retrieval and so on. 
With the rapid development of deep learning techniques, intensive research interests have been paid for this  topic. 


\begin{figure}[htbp]
	\centering {\includegraphics[height=2in, width=3.5in]{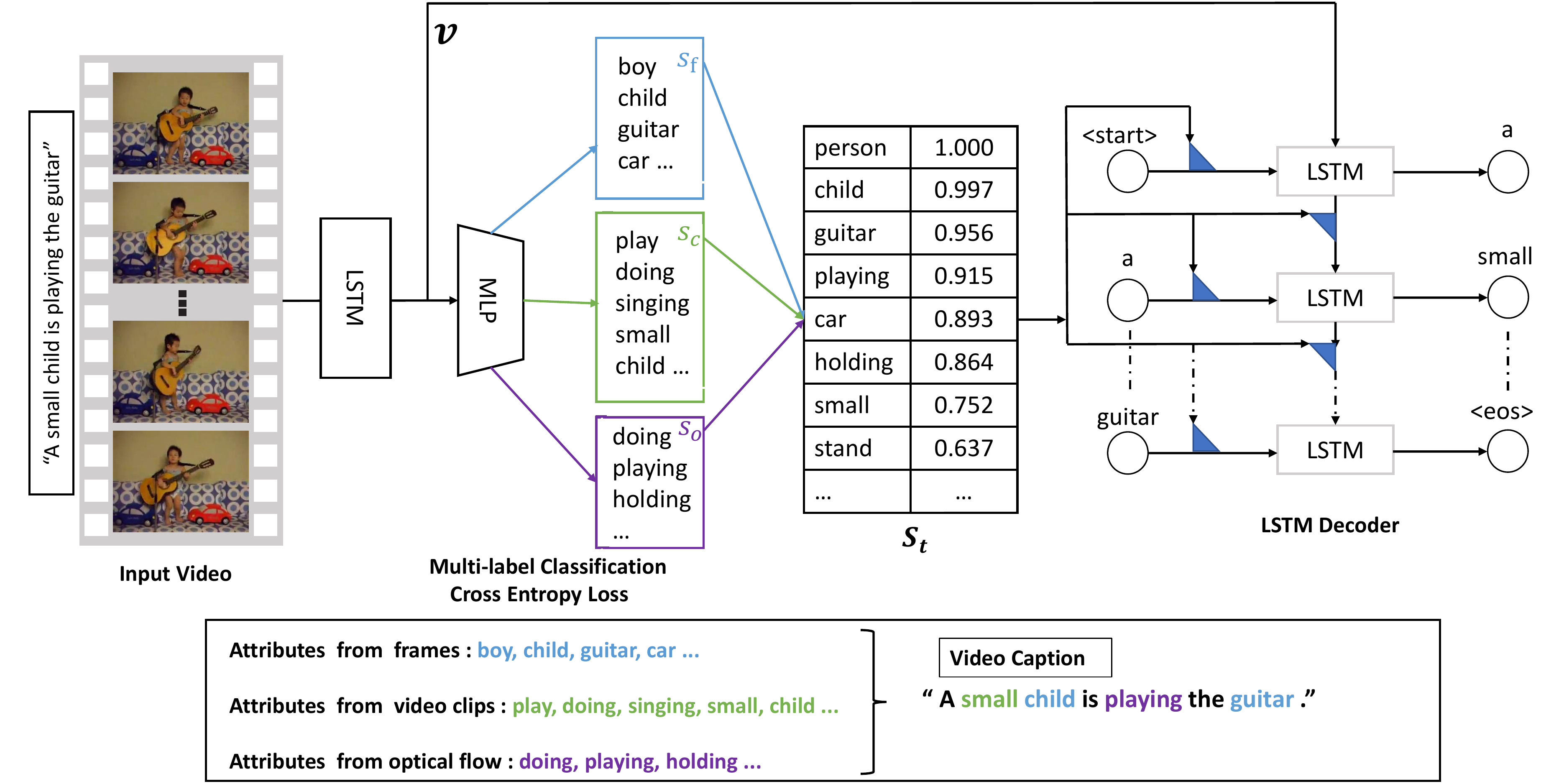}}
	\vspace{-0.25in}
	\caption{An example of video description generation, 
		the attributes are learned from multimodal features. Words in blue color, green color, purple color correspond to visual frame, video clip and optical flow respectively.
		The number next to a semantic attribute is the probability of the corresponding semantic 
		presented in this video.}
	\label{fig:pic1}
	\vspace{-0.1in}
\end{figure}

Inspired by the success of the encoder-decoder framework in machine translation \cite{b2}, 
most work on video \hyphenation{captioning} captioning employs a convolutional neural network (CNN) as an encoder to obtain a fixed length vector representation, then a recurrent neural network (RNN) is employed as a decoder to generate a caption. 
These CNN plus RNN-based learning \hyphenation{approaches} approaches  translate directly from video representation to language without taking any high-level semantic concepts into account.  
Recent work \cite{b3,b4} improves the performance of visual captioning by adding explicit high-level semantic attributes of the image/video. However, most of methods only use attributes learnt from single modality,  
it is possible not to understand the key actors, objects and their interaction in the scene adequately. 
Some methods \cite{b3,b5,b6} integrating semantic attributes into the RNN-based caption generation process are mainly constrained initialization of the first step of the RNN or through soft attention.  
They represent the semantic vector as a whole, so couldn't mine the meanings of individual words to generate the caption. 

In order to solve the aforementioned limitations, we propose a novel deep architecture, named  \textit{Multimodal Semantic Attention Network} (MSAN), which takes advantage of incorporating multimodal semantic attributes into sequence learning for video captioning.
We capture three modalities of features and their corresponding semantic attributes to  represent videos. 
Take the given video in Figure \ref{fig:pic1} as an example, the semantic attributes learnt from image frames often depict static objects and scenes (e.g., ``boy", ``child" and ``guitar"), while the semantics extracted from optical flow frames and video clips often convey temporal dynamic motions (e.g., ``playing", ``doing" , and ``holding"). 
This has made the attributes mined from different modalities complementary to each other for the sentence generation (e.g., ``a small child is playing the guitar"). 
Meanwhile, we investigate how the attributes from the three sources can be leveraged to enhance video captioning, 
and propose a new fusion method that extends each weight matrix of the conventional RNN to an ensemble of attributes-dependent weight matrices. 
Considering that different semantic attributes have a different impact on sentence generation, 
we adopt an attention-based fusion strategy to let the model selectively focus on different  semantic information parts of the video each time it produces a word. 
In general,  
our main contributions can be summarized as follows:

\noindent 1) We propose a new encoder-decoder network exploiting the multimodal semantic attributes for video captioning.

\noindent 2) We add a multimodal semantic classification loss to our deep neural network, it is optimized with video captioning loss simultaneously.

\noindent 3) We incorporate the attention mechanism into the LSTM decoder to automatically focus on different semantic attributes for caption generation.

\noindent 4) We perform comprehensive evaluations on two popular video captioning benchmarks, demonstrating that  proposed  method outperforms previous state-of-the-art 
 approaches.

\vspace{-0.2in}
\section{Related work}
\vspace{-0.1in}
In this section, we briefly review the related works in three aspects: video captioning, video captioning with attention, and video captioning with semantic attributes.

\textbf{Video Captioning}. The research on video captioning mainly includes two different dimensions: template-based language methods \cite{b7,b36} and sequence learning approaches \cite{b9,b11,b14}. 
The former predefines a set of templates for sentence generation following specific grammar rules,
Li et al. \cite{b36} extract the phrases related to detected objects, attributes and their relationships for video captioning. 
Obviously, this approach highly depends on the template of sentences and always generates sentences with the same syntactical structure. 
Different from template-based language methods, sequence learning method  
directly translates the video content into a sentence, 
and it learns the probability distribution in the common space of visual content and textual sentence. 

\textbf{Video captioning with attention}. 
Attention mechanisms have been used to boost the network's ability to select the relevant features from the corresponding parts of the input. 
In video captioning, 
people do not describe everything in a video, instead they tend to talk more about semantically important regions and objects. \cite{b22} utilizes a  spatial attention-based mechanism to learn where to focus in the image. 
This work is followed by \cite{b12} which introduces a temporal attention-based mechanism module to exploit temporal structure for video captioning. 
Recently, a new use of multimodal fusion attention is proposed to fuse information across different modalities in \cite{b23}.

\textbf{Video captioning with semantic attributes}. 
Attributes are properties observed in visual content with rich semantic cues, recent work \cite{b3,b4,b5,b6,b9,b10} show that adding  high-level semantic concepts can further improve visual captioning. Multiple Instance Learning is used as an \hyphenation{attribute}attribute detector in \cite{b4} and then generates sentence based on the outputs of attribute detector. \cite{b5} applies retrieved
sentences as additional semantic information to guide the LSTM when generating captions. 
A transfer unit is designed in \cite{b9} to transfer the information of semantic attributes from image and video to boost video captioning. In \cite{b6}, a model of semantic attention is proposed which selectively attends to semantic concepts through a soft attention mechanism, 
in which a score is assigned to each detected attribute based on its relevance with the previous predicted word, but our work regards the whole video as a label to learning semantic probability distribution. 
Generally, our work is different from most of the aforementioned sequence learning models, which 
only uses single modality, but our work extracts semantic attributes by multimodal features. Meanwhile,
our work adds a multimodal semantic classification loss to deep neural network, it is optimized with video captioning loss simultaneously. 

\begin{figure*}[htbp]
	\vspace{-0.3in}
	\centering {\includegraphics[height=2.5in, width=6.4in]{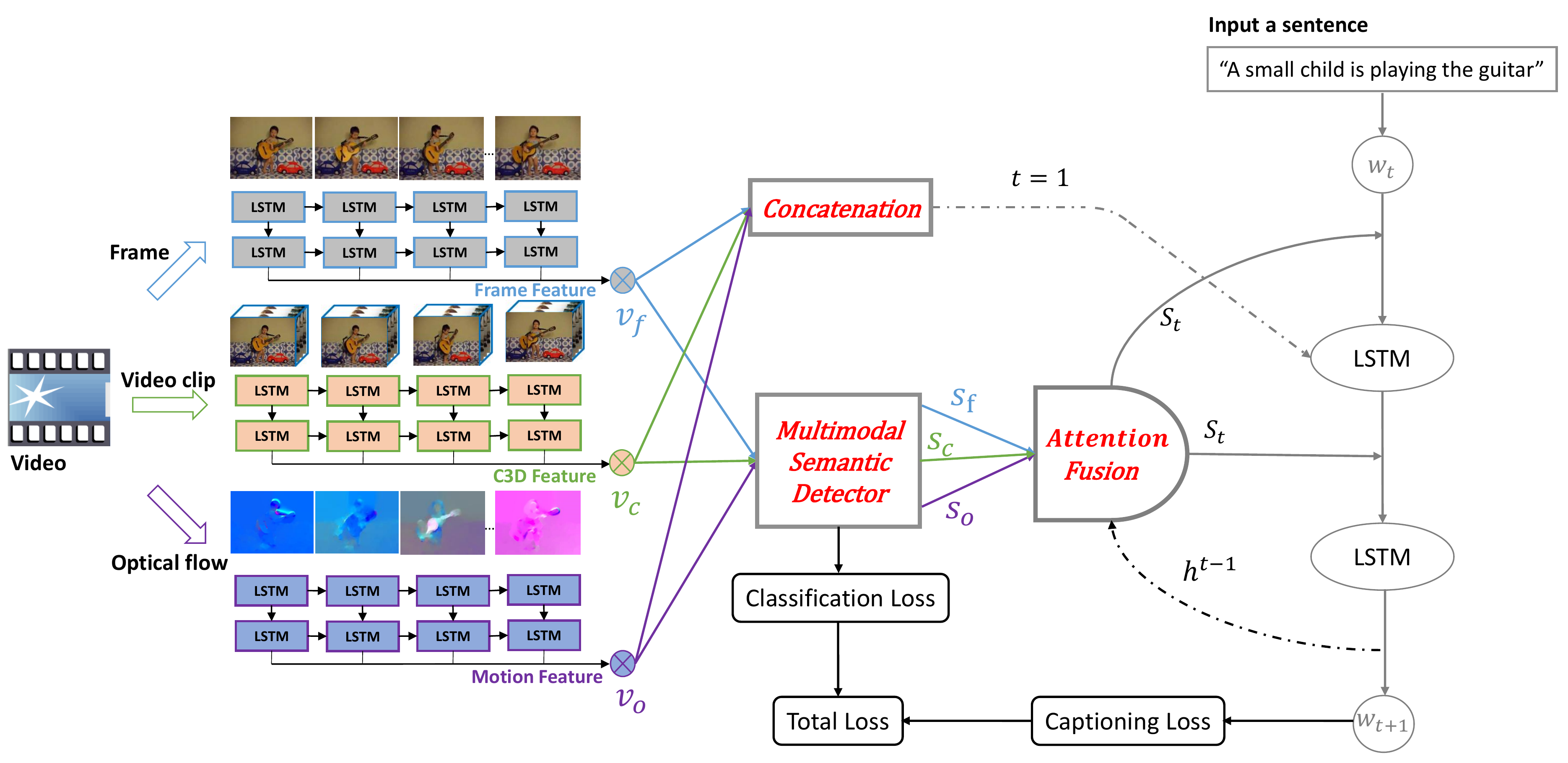}}
	\vspace{-0.2in}
	\caption{The overall architecture of MSAN. 
	In the encoder phase, three LSTM models are used to encode features ($\bm{v}_f,\bm{v}_c,\bm{v}_o$) from different modalities (video frames, video clips and optical flow)  separately, then semantic attributes ($\bm{s}_f,\bm{s}_c,\bm{s}_o$) are learned through a semantic detector. In the decoder phase, video feature vector $\bm{v}$ is obtained through a concatenation, 	
		multimodal semantic attributes $S_t$ is used to help better capture the key semantic clues in videos. 
	}
	\label{fig:pic2}
\end{figure*}

\begin{figure}[htbp]
	\vspace{-0.22in}
	\centering {\includegraphics[width=0.5\textwidth]{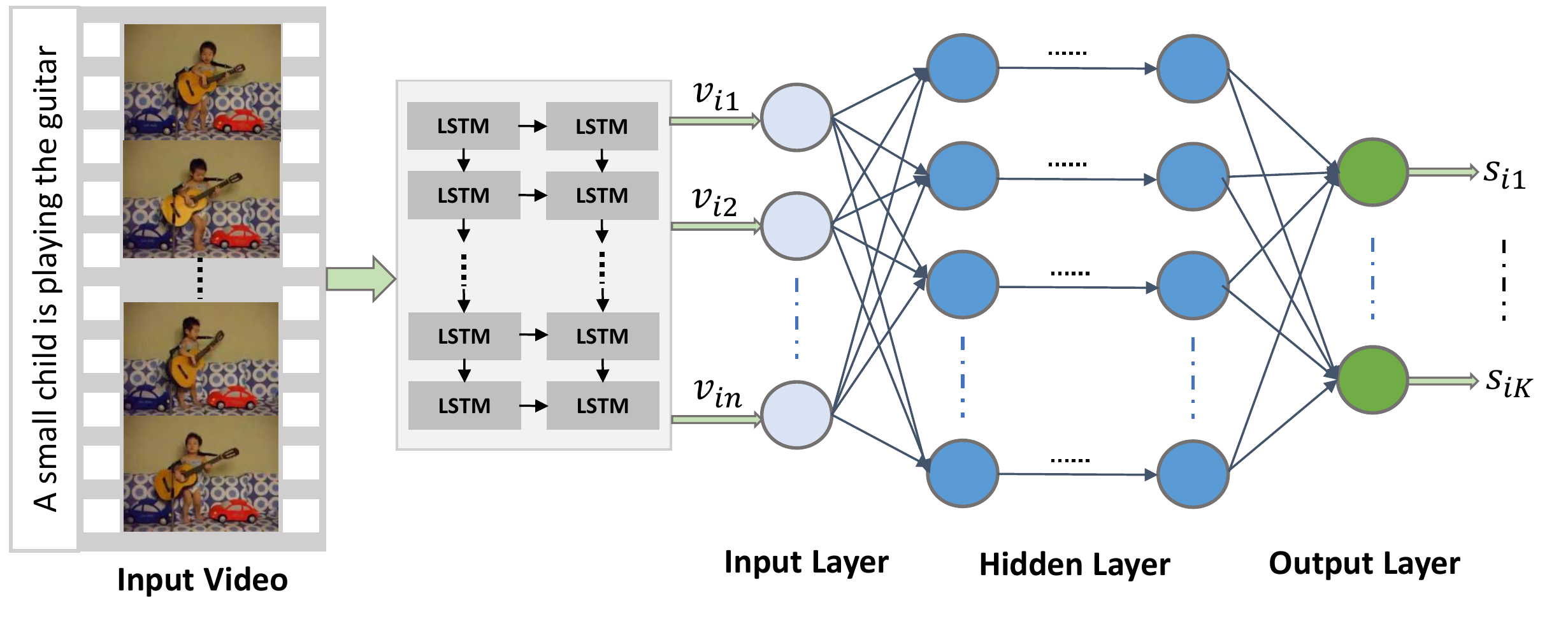}}
	\vspace{-0.32in}
	\caption{
		Semantic attribute detector. 
	}
	\label{fig3}
	\vspace{-0.22in}
\end{figure}

\vspace{-0.1in}
\section{Multimodal Semantic Attention Network}
\vspace{-0.1in}

The overall architecture of MSAN is shown in Figure \ref{fig:pic2},
it mainly consists of two modules: LSTM-based encoder with multimodal semantic attributes and  attention-based LSTM decoder.  
It is trained end-to-end with a joint loss on all aforementioned targets. 

\vspace{-0.1in}
\subsection{Encoder with Multimodal Semantic Attributes}\label{sec:seman_detect}

In the encoding phase, 
we respectively extract three CNN feature sequences from frames, video clips, and optical flow frames for each video $V$, then we employ three two-layer LSTM to model the obtained feature sequences  and obtain the corresponding video representations $\bm{v}_f$, $\bm{v}_c$, and $\bm{v}_o$. The video  $V$ is finally represented as $\bm{v}$ by concatenating the three video representations. 
Moreover, we add a multimodal semantic attribute detector to exploit high-level semantic attributes for further improving video captioning. 

Figure \ref{fig3} shows the semantic attribute detector architecture.   
Once obtained visual features $\bm{v}_f$, $\bm{v}_c$, and $\bm{v}_o$, 
we adopt the multi-label classification approach to learn semantic attributes $\bm{s}_f$, $\bm{s}_c$ and $\bm{s}_o$, which represent the probability distributions over the high-level attributes for video.
Specifically, 
we first sort all words extracted from training and validation sets by their frequency, then remove some function words (such as ``a'' ``the") 
and finally selecte the top K words including verbs, nouns, adjs as semantic attributes. 
Suppose there are $N$ training examples, and $\bm{y}_i=[y_{i1},\dots,y_{ik}]$ is the label vector of the $i$-th video,  where $y_{ik}=1$ if its caption includes the word $k$, and $y_{ik}=0$ otherwise. Let $\bm{v}_i$ represent the $i$-th video feature vector obtained from multiple LSTM layers. Then we employ a MLP to learn a function $f(\cdot): R^m \rightarrow R^K$ by the training examples $\{\bm{v}_i, \bm{y}_i\}$, where $m$ is the number of dimensions for input and the number $K$ of dimensions for output is equal to the number of semantic attributes. Let $\bm{s}_i = [s_{i1},\dots, s_{iK}]$ be the predicted label vector for the $i$-th video, which is namely the semantic attribute distribution we want to learn. The multi-label classification loss (loss$_1$) is defined as follows,
\vspace{-0.15in}
\begin{align}\label{eq:detect}
\begin{split}
loss_1 &= \frac{1}{N}\sum_{i=1}^{N}\sum_{k=1}^{K}(y_{ik}\log s_{ik} + (1-y_{ik})\log (1-s_{ik}))  \\
       &  + \alpha ||W^{encoder}||_2^2,
\end{split}
\end{align}
\vspace{-0.25in} \\
where $W^{encoder}$ is the parameters of the encoder model, $\bm{s_i} = \sigma(f(\bm{v}_i))$ is a $K$-dimensional vector, $\sigma(\cdot)$ is  logistic sigmoid function and $f(\cdot)$ is implemented as a multilayer perceptron. 

\vspace{-0.15in}
\subsection{Attention-based LSTM Decoder}
\vspace{-0.05in}
In the decoder phase, an attention LSTM model is proposed to generate the textual sentence by combining both the visual features and semantic features. 
Given a video, the goal of video captioning is to output a textual sentence $Y$, where $Y=\{\bm{w}_1,\bm{w}_2,\dots,\bm{w}_t,\dots,\bm{w}_{N_s}\}$ consists of $N_s$ words. 
The video sentence generation problem can be formulated by minimizing the following captionign loss (loss$_2$) function as
\vspace{-0.1in}
\begin{align}\label{eq:energy}
\begin{split}
loss_2 &=-\log P(Y|\bm{v},\bm{s}_f,\bm{s}_c,\bm{s}_o) \\
&=-\sum_{t=1}^{N_s}\log P(\bm{w}_t|\bm{v},\bm{s}_f,\bm{s}_c,\bm{s}_o,\bm{w}_{0\sim{t-1}})
\end{split}
\end{align}
\vspace{-0.2in} \\
which is the negative $\log$ probability of the correct textual sentence given the video and the detected multimodal semantic attributes. 
%
In training phrase, the total loss being optimized, is sum of  $loss_1 + loss_2$.  
By minimizing this total loss, the contextual relationship  in the sentence can be guaranteed given video feature and its learnt multimodal semantic attributes.

As discussed in Equation (\ref{eq:energy}), 
we employ the LSTM-based decoder to generate a sentence for each video. Given input word $\bm{w}_t$, last  hidden  state $\bm{h}_{t-1}$, and last memory cell $\bm{c}_{t-1}$, the LSTM is updated for time step $t$ as following :
\vspace{-0.1in}
\begin{align}
\bm{i}_t &= \sigma(W_{i}\bm{w}_{t}+ U_{hi}\bm{h}_{t-1} +\bm{z} ),\label{eq:4}\\
\bm{f}_t &= \sigma(W_{f}\bm{w}_{t}+ U_{hf}\bm{h}_{t-1} +\bm{z} ),\label{eq:5}\\
\bm{o}_t &= \sigma(W_{o}\bm{w}_{t}+ U_{ho}\bm{h}_{t-1} +\bm{z} ),\label{eq:6}\\
\bm{\tilde{c}}_t &= \tanh(W_{c}\bm{w}_{t}+ U_{hc}\bm{h}_{t-1} +\bm{z} ),\label{eq:7}\\
\bm{c}_t &= \bm{i}_t\odot \bm{\tilde{c}}_t + \bm{f}_t\odot \bm{c}_{t-1},\\
\bm{h}_t &= \bm{o}_t\odot tanh(\bm{c}_t),\\
\bm{z} &= 1(t=1)\cdot C\bm{v}
\end{align}
\vspace{-0.3in} \\
Let $\ast$ denote one subscript among $\bm{i},\bm{f},\bm{o}$, and $\bm{c}$ in the above equations. Here $W_{\ast}, U_{h\ast}$ and $C$ represent the weight matrices,
$\bm{i}_t, \bm{f}_t, \bm{o}_t, \bm{c}_t, \bm{\tilde{c}}_t$ represent the states of input gate,  forget gate,  output gate,  memory cell  and
squashed input, respectively, at time $t$.
$\tanh(\cdot)$ is hyperbolic tangent function,
and $1(t = 1)$ is an indicator function,
which represents that video feature vector $\bm{v}$ is fed into the LSTM at the beginning. For Simplify, bias terms are omitted throughout the paper.

To better exploit the complementary information from multiple semantic attributes, we propose to combine them to compute  weight matrices $W_{\ast}, U_{h\ast}$ by using a attention unit.
We extend each weight matrix of the conventional LSTM to an ensemble of $K$ attributes-dependent weight matrices for mining the meanings of individual words to generate the caption. Namely, we replace $W_{\ast} / U_{h\ast}$ with $W_{\ast}(S_{t}) /U_{h\ast}(S_{t})$ for each $\ast \in \{\bm{i},\bm{f},\bm{o},\bm{c}\}$, where $S_t \in R^K$ is a multimodal semantic attributes vector.
Specifically, we define two weight tensors $W_\tau \in R^{n_h\times n_x\times K}$ and $U_\tau \in R^{n_h\times n_h\times K}$, where $n_h$ is the number of hidden units and $n_x$ is the dimension of word embedding. $W_{\ast}(S_{t}) \in R^{n_h\times n_x}$ and $U_{h\ast}(S_t) \in R^{n_h\times n_h}$ can be written as:
\vspace{-0.25in}
\begin{align}
W_{\ast}(S_t) &= \sum_{k=1}^{K}S_t[k]W_{\tau}[k],\label{eq:15}  \\
U_{h\ast}(S_t) &= \sum_{k=1}^{K}S_t[k]U_{\tau}[k],\label{eq:16}
\end{align}
\vspace{-0.2in} \\
where $W_{\tau}[k]$ and $U_{\tau}[k]$ represent the $k$-th 2D slice of $W_{\tau}$ and $U_{\tau}$ respectively, which are associated with probability $S_t[k]$, and
$S_t[k]$ is the $k$-th element in $S_t$. It implicitly specifies $K$ LSTMs in total.  
In order to combine $K$ LSTMs, we propose to learn an attention-based  multimodal semantic attributes vector $S_t$  at each time step $t$. It is defined as
\vspace{-0.15in}
\begin{align}
S_t=\sum_{i=1}^{l}a_{ti}\bm{s}_i, 
\end{align}
\vspace{-0.2in} \\
where $l=3$ represents that we have learned three semantic attributes $(\bm{s}_f,\bm{s}_c, \bm{s}_o)$.
The attention weight $a_{ti}$ reflects the relevance of the $i$-th semantical attribute in the input video given all the previously generated words.
Hence, we design an attention unit to calculate $S_t$ that takes both the previous hidden state $\bm{h}_{t-1}$, 
and the $i$-th semantic attribute vector as input and returns the unnormalized relevance score $e_{ti}$ :
\vspace{-0.1in}
\begin{align}
e_{ti} = \bm{w}^T\tanh (W_a\bm{h}_{t-1} + U_a\bm{s}_i), 
\end{align}
\vspace{-0.25in}\\
where ${W_a,U_a}$ and $\bm{w}$ are the parameters that are estimated together with all the other parameters in networks.
Once the relevance score $e_{ti}$ 
are computed, we normalize them
to obtain the $a_{ti}:$
\vspace{-0.2in}
\begin{align}
a_{ti} = \frac{\exp \{e_{ti}\}}{\sum_{j=1}^{l}\exp \{e_{tj}\} }.
\end{align}
\vspace{-0.15in} \\
It can be seen that for different time step $t$, the semantic attributes $S_t$ are different, which makes the model selectively focus on different semantic information parts of the video at each time when it produces a word.


Because of training such a model defined in (\ref{eq:15}) and (\ref{eq:16}) is the same as jointly training an ensemble of $K$ LSTMs, 
though appealing, the number of parameters is proportional to $K$, which is unrealistic for large $K$. We factorize $W_{\ast}(S_t)$ and $U_{h\ast}(S_t)$ define in (\ref{eq:15}) and (\ref{eq:16}) as:
\vspace{-0.12in}
\begin{align}
W_{\ast}(S_t) &= W_a\cdot diag(W_bS_t)\cdot W_c ,\label{eq:17}\\
U_{h\ast}(S_t) &= U_a\cdot diag(U_bS_t)\cdot U_c ,\label{eq:18}
\end{align}
\vspace{-0.25in} \\
where $W_a \in R^{n_h\times n_f} $, $W_b \in R^{n_f\times K} $, and $W_c \in R^{n_f\times n_x} $. Similarly,
$U_a \in R^{n_h\times n_f} $, $U_b \in R^{n_f\times K} $, and $U_c \in R^{n_f\times n_h} $, $n_f$ is the number
of factors which are the concept of factorization.
Substituting (\ref{eq:17}) and (\ref{eq:18}) into (\ref{eq:4}), we can obtain attention-based LSTM decoder as:
\vspace{-0.15in}
\begin{align}
\bm{i}_t&=\sigma(W_{a}\bm{\hat{w}}_{t}+ U_{hi}\bm{\hat{h}}_{t-1} +\bm{z} ), \label{eq:ii}\\
\bm{\hat{w}}_t &= W_bS_t\odot W_c\bm{w}_t,\label{eq:20} \\
\bm{\hat{h}}_{t-1} &= U_bS_t\odot U_c\bm{h}_{t-1},\label{eq:21}
\end{align}

\vspace{-0.15in}
{\noindent}where $\odot$ represents the element-wise multiply operator. In (\ref{eq:17}) and (\ref{eq:18}), $W_a$ and $U_a$ are shared among all the captions, which can effectively capture common linguistic patterns. Meanwhile, the diagonal terms  ($W_bS_t$ and $U_bS_t$) are captured by $S_t$, which accounts for the specific semantic attributes in $\bm{i}_t$  of the video under test.
$\bm{f}_t,\bm{o}_t, \bm{c}_t$ are obtained the same as equation (\ref{eq:ii}). 
Hence, via the decomposition in (\ref{eq:17}) and (\ref{eq:18}), our network effectively learn an ensemble of $K$ sets of LSTM parameters, one key word in $S_t$ corresponds to a set of parameters in one LSTM.
By sharing $W_a$ and $W_c$ when composing each member of the ensemble, we can remedy this problem of a large $K$. 

\vspace{-0.15in}
\section{Experiments}\label{sec:experiment}
\vspace{-0.1in}
We evaluate the proposed MSAN model on two standard video captioning benchmarks: MSVD \cite{b24} and MSR-VTT \cite{b25}.
The MSVD dataset consists of 1,967 short videos, 
we follow the setting used in prior works \cite{b7,b14}, taking 1,200 videos for training, 100 ones for validation and 670 ones for test in our experiments. 
The MSR-VTT dataset 
contains 10,000 video clips in 20 well-defned categories,  
we use the data split defined in \cite{b25} in our experiments: 6,513 videos for training, 497 ones for validation, and 2,990 ones for test.

\vspace{-0.12in}
\subsection{Experimental Settings}

For training, all the parameters in the MSAN are initialized from a uniform distribution in [-0.05,0.05], all bias are initialized to zero.  We set both the number of hidden units and the number of factors to 512, 
word embedding vectors are initialized with the publicly available word2vec vectors.  
The maximum number of epochs for all the two datasets is 20, gradients are clipped if the norm of the parameter vector exceeds 5. We use dropout and early stopping on validation sets, and the Adam algorithm  with learning rate $1\times 10^{-4}$ is utilized for optimization. 

In test stage, we adopt the beam search strategy for caption generation and set beam size to 5.
For quantitative evaluation of our proposed models, we adopt three common metrics in video captioning: BLEU@N \cite{b31}, METEOR \cite{b32}, and CIDEr-D \cite{b33}. 
All metrics are computed by \hyphenation{using}using the codes   released by Microsoft COCO Evaluation Server \cite{b34}.

\vspace{-0.1in}
\subsection{Quantitative Analysis}\label{sub:Quantitative}
At first, we aim to evaluate the effect of using different features and their combination specifically. 
In our MSAN model, we test six variants using six different semantic attributes, 
and concatenate $\bm{v}_f,\bm{v}_c,\bm{v}_o$ as final video  feature $\bm{v}$, which is fed into the LSTM decoder at the beginning.  The ``MSAN$_f$" only uses the semantic attributes $\bm{s}_f$, and the ``MSAN$_{f+o}$"  uses two semantic attributes $\bm{s}_f$ and $\bm{s}_o$. The same denotations are used for other four models. 
$f,o,c$ denote semantic attrbutes $\bm{s}_f,\bm{s}_o$  and $\bm{s}_c$ respectively.   
Table \ref{table:MSVD} shows the performances of different models on the MSVD dataset.  
\vspace{-0.15in}
\begin{table}[htbp]\small
	\centering
	\caption{METEOR, CIDEr-D, and BLEU@N scores of our MSAN and other state-of-the-art methods on the MSVD dataset. 
		All values are reported as percentage (\%).  }
	\label{table:MSVD}
	
	\resizebox{3.4in}{!}{
		
		\begin{tabular}{l|c|c|c|c|c|c}\hline
			~~\textbf{Model}&~~\textbf{METEOR}~~&~~\textbf{CIDEr-D}~~&~~\textbf{BLEU@1}~~&~~\textbf{BLEU@2}~~&~~\textbf{BLEU@3}~~&~~\textbf{BLEU@4}~~
			\\ \hline\hline
			~~LSTM \cite{b14} & 29.1 & - &- & - & - & 33.3  \\
			~~S2VT \cite{b11} & 29.8 &- &- & - & - & -  \\
			~~TA \cite{b12}     & 29.6 & 51.7 &80.0 & 64.7 & 52.6 & 41.9  \\
			~~LSTM-E \cite{b13} & 31.0 & - &78.8 & 66.0 & 55.4 & 45.3  \\
			~~GRU-RCN \cite{b16} & 31.6 & 68.0 &- & - & - & 43.3                \\
			~~h-RNN  \cite{b17} & 32.6 & 65.8 &81.5 & 70.4 & 60.4 & 49.9 \\
			~~HRNE \cite{b15} & 33.1 & - & 79.2 & 66.3 & 55.1 & 43.8 \\
			~~LSTM-TSA$_{IV}$ \cite{b9} &33.5 & 74.0 & 82.8 & 72.0 & 62.8 & 52.8 \\ \hline
			~~\textbf{MSAN$_{f}$} &33.5 & 77.7 & 80.9 & 69.7 & 60.5 & 51.1 \\
			~~\textbf{MSAN$_{o}$} &32.9 & 71.7 & 82.1 & 70.7 & 61.1 & 50.5 \\
			~~\textbf{MSAN$_{c}$} &33.1 & 69.2 & 81.5 & 69.3 & 59.5 & 50.3 \\
			~~\textbf{MSAN$_{f+o}$} &34.0 & 75.7 & 82.1 & 72.3 & 62.1 & 52.5 \\
			~~\textbf{MSAN$_{f+c}$} &34.3 & 78.5 & 82.6 & 73.3 & 63.2 & 53.2 \\
			~~\textbf{MSAN$_{o+c}$} &33.6 & 71.9 & 82.3 & 70.8 & 61.2 & 51.0 \\
			~~\textbf{MSAN$_{f+o+c}$} &\textbf{35.3} &\textbf{79.6} &\textbf{84.1} & \textbf{75.2} &\textbf{69.1}  & \textbf{56.4} \\ \hline
			
		\end{tabular}
		\vspace{-0.1in}
		
	}
	
	\vspace{-0.2in}
\end{table}

From Table \ref{table:MSVD},  we can conclude the following points. 1) The results across six evaluation metrics consistently indicate that our proposed MSAN$_{f+o+c}$ achieves the best performance than all the state-of-the-art methods. In particular, the METEOR and CIDEr-D of our MSAN$_{f+o+c}$ can achieve 35.3\% and 79.6\% which are to date the highest performance reported on MSVD dataset.
2) Incorporating different attributes to MSAN, such as MSAN$_f$, MSAN$_{f+o}$, and MSAN$_{f+o+c}$, 
the results across six evaluation metrics are gradually increasing, which indicates that visual representations are augmented with multimodal semantic attributes and thus do benefit the learning of video sentence generation. 
Notably, MSAN$_{f+o+c}$ 
improves performance of MSAN$_f$ and MSAN$_{f+o}$, which demonstrates the advantage of leveraging the learnt multimodal semantic attributes for boosting video captioning.
3) Even only using one kind of semantic attributes $s_f$, 
the MSAN$_{f}$ achieves  competitive results with LSTM-TSA$_{IV}$ across different evaluation metrics, which proves the effectiveness of the proposed MSAN framework. Namely. 

\vspace{-0.15in}
\begin{table}[htbp]\small
	\centering
	\caption{The scores of our MSAN$_{f}$ and other  methods that use single semantic information for caption generation on MSVD dataset. All values are reported as percentage (\%).}
	\label{table:MSVD1}	
	\resizebox{3.4in}{!}{
		
		\begin{tabular}{l|c|c|c|c|c|c}\hline
			~~\textbf{Model}&~~\textbf{METEOR}~~&~~\textbf{CIDEr-D}~~&~~\textbf{BLEU@1}~~&~~\textbf{BLEU@2}~~&~~\textbf{BLEU@3}~~&~~\textbf{BLEU@4}~~
			\\ \hline\hline
			~~LSTM-$v$  & 31.0 & 64.0 &- & - & - & 44.8  \\
			~~LSTM-$vf$  & 31.6 &64.7 &- & - & - & 47.5  \\
			~~LSTM-$vf_2$      & 32.5 & 70.6 &73.6 & 59.3 & 48.3 & 46.9  \\ \hline			
			~~\textbf{MSAN$_{f}$} &\textbf{33.5} & \textbf{77.7} & \textbf{80.9} & \textbf{69.7} & \textbf{60.5} & \textbf{51.1} \\ \hline
			
		\end{tabular}

	}	
	\vspace{-0.15in}
\end{table}

Secondly, we further compare our MSAN$_{f}$ model with other three models that adopt different attributes fusion strategy for caption generation. For fair comparison,  all these methods are based on single semantic attributes $\bm{s}_f$, and use the same video feature $\bm{v}$. In the denotations of ``LSTM-$\bm{v}$/LSTM-$\bm{v}f$", $\bm{v}$ denotes video feature vector, $f$ denotes the semantic attributes vector $\bm{s}_f$ that derives from video RGB frames, and $\bm{v}f$ denotes the concatenation of $\bm{v}$ and $\bm{s}_f$, which is fed into a standard LSTM decoder only at the initial time step.  
Actually, the LSTM-$\bm{v}$ \cite{b14} model can be treated as a baseline architecture, which doesn't use semantic attributes.  LSTM-$\bm{v}f$ is the model proposed in \cite{b3}, which uses the concatenation of  $\bm{v}$ and $\bm{s}_f$ to feed into a standard LSTM decoder. 
In LSTM-$\bm{v}f_2$, the video feature vector is sent to a standard LSTM decoder at the first time step, while the semantic vector is sent to the LSTM decoder at every time step in addition to the input word. This model is similar to \cite{b6} without using semantic attention.
MSAN$_{f}$ is one of our model. 

\begin{table}\small
	\centering
	\caption{The scores of our MSAN and other methods on MSR-VTT dataset,  all values are reported as percentage (\%).
	}
	\label{table:MSR}
	
	\resizebox{3.4in}{!}{
		
		\begin{tabular}{l|c|c|c|c|c|c}\hline
			~~\textbf{Model}&~~\textbf{METEOR}~~&~~\textbf{CIDEr-D}~~&~~\textbf{BLEU@1}~~&~~\textbf{BLEU@2}~~&~~\textbf{BLEU@3}~~&~~\textbf{BLEU@4}~~
			\\ \hline\hline
			~~LSTM \cite{b14} & 23.7 & 35.0 &- & - & - & 30.4  \\
			~~S2VT \cite{b11} & 25.7 &35.2 &- & - & - & 31.4  \\
			~~TA \cite{b12}     & 25.0 & 37.1 &80.0 & 64.7 & 52.6 & 28.5  \\ 
			~~$M^3$-VC \cite{b35}     & 26.6 & - &73.6 & 59.3 & 48.3 & 38.1  \\ \hline
			
			~~\textbf{MSAN$_{f}$} &27.3 & 45.2 & 79.4 & 62.1 & 53.4 & 42.5 \\
			~~\textbf{MSAN$_{o}$} &26.8 & 39.6 & 76.3 & 59.7 & 51.5 & 39.8 \\
			~~\textbf{MSAN$_{c}$} &27.1 & 43.7 & 78.6 & 61.3 & 52.9 & 40.7 \\
			~~\textbf{MSAN$_{f+o}$} &28.2 & 47.2 & 81.3 & 66.5 & 55.8 & 43.4 \\
			~~\textbf{MSAN$_{f+c}$} &28.6 & 48.3 & 81.9 & 65.7 & 54.2 & 44.6 \\
			~~\textbf{MSAN$_{o+c}$} &27.9 & 46.9 & 80.5 & 62.9 & 54.7 & 42.7\\
			~~\textbf{MSAN$_{f+o+c}$} &\textbf{29.5} & \textbf{52.4} & \textbf{83.4} & \textbf{68.4} & \textbf{58.9} & \textbf{46.8} \\ \hline
			
		\end{tabular}
		
	}		
	\vspace{-0.3in}	
\end{table}

The results of above four methods are listed in Table \ref{table:MSVD1}. It can be seen that  our  MSAN$_{f}$ achieves the best performance than all other models, which proves that our semantic  
fusion method can effectively integrate semantic attributes for captioning. Especially, the improvement of our MSAN$_{f}$ compared with 
LSTM-$v$ is huge, which demonstrates it is very necessary to combine the semantic attributes. The performance of of our MSAN$_{f}$ is higher than that of LSTM-$vf_2$, which demonstrate that using attention mechanism to obtain different semantic attribute vectors at different time steps fed into LSTM can further boost sentence generation compared with using the same semantic attribute vector at different time steps. Therefore, it is useful to combine the attention mechanism to incorporate semantic attributes into LSTM decoder. 
Finally, we test six variants of our MSAN model and other methods on the MSR-VTT dataset, 
which concludes the consistent results with the MSVD dataset in Table \ref{table:MSR}.
Our models almost outperform    
all competing methods across all evaluation metrics, especially our MSAN$_{f+o+c}$ achieves an improvement by a substantial margin.

\vspace{-0.2in}
\subsection{Qualitative Analysis}
\begin{figure}\small
	
	\centering {\includegraphics[width=0.48\textwidth]{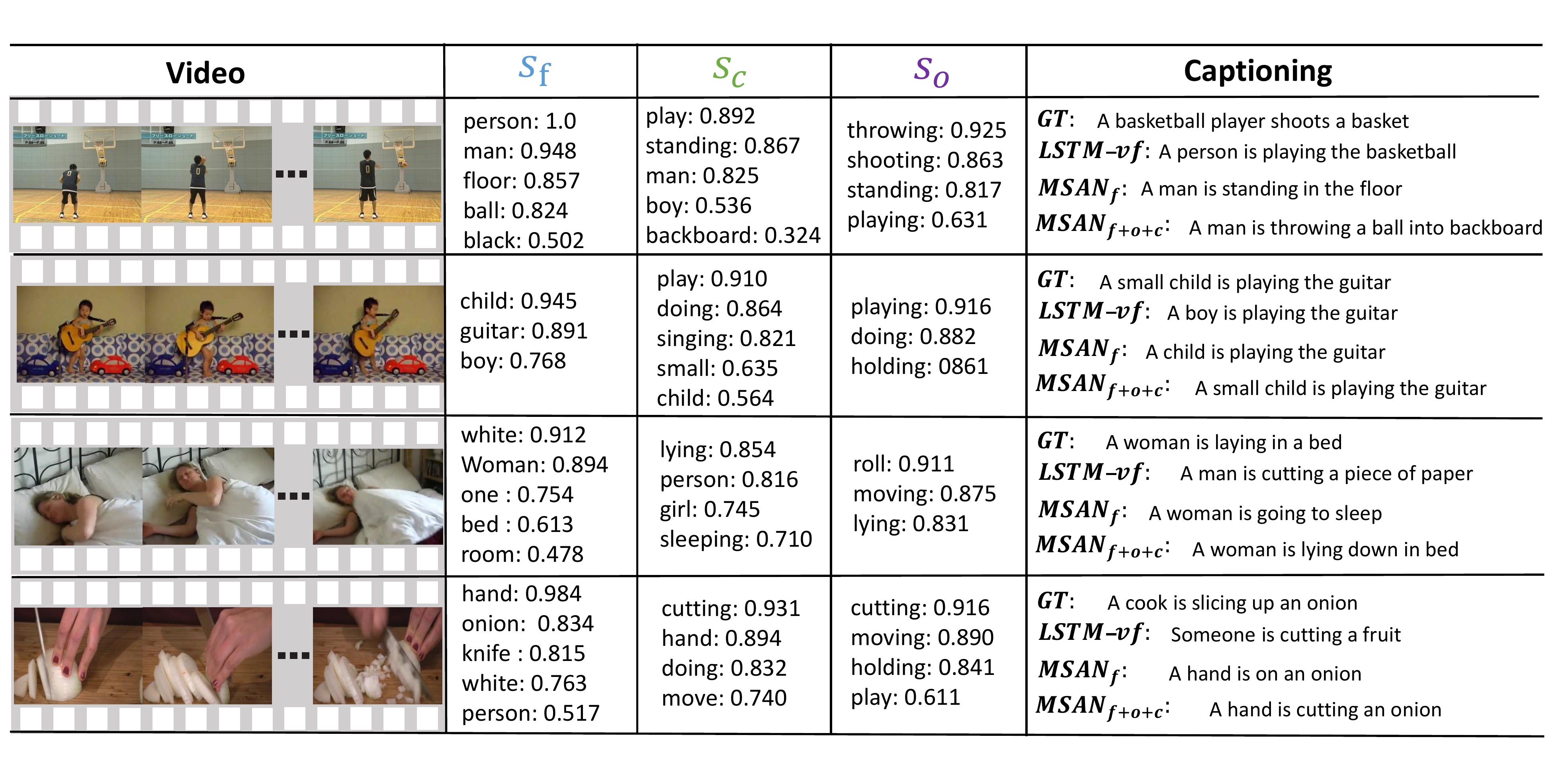}}
	\vspace{-0.45in}
	\caption{
		The output sentences are generated on MSVD dataset. 
		GT: Ground Truth, LSTM-$vf$, MSAN$_{f}$ and  MSAN$_{f+o+c}$.}
	\label{fig:result}
	\vspace{-0.25in}
\end{figure}

Figure \ref{fig:result} shows four examples, including 
ground truth sentences (GT) and the sentences generated by three approaches. 
From these example results, it is obviously to see that our models can generate somewhat relevant and logically correct sentences. 
For instance, compared to the subject term ``a person'' and the verb term ``playing'' in the sentence generated by LSTM-$vf$ for the first video, ``a man'' and ``throwing'' in our MSAN$_{f+o+c}$ are more relevant to the video content, since the word ``man'' and ``throwing'' predicted as one attribute from different modalities which complement to each other. 

Moreover, compared with  MSAN$_{f}$, MSAN$_{f+o+c}$ can generate more descriptive sentences by enriching the semantics with attributes. For instance, with the verb term ``lying'' learnt from $\bm{s}_c$ and $\bm{s}_o$, the generated sentence ``A woman is lying down in bed'' of the third video depicts the video content more comprehensive. This confirms that video captioning is improved by leveraging complementary multimodal semantic attributes learnt from videos.

\vspace{-0.15in}
\section{Conclusions}
\vspace{-0.1in}
We have proposed a MSAN framework which explores both video representations and multimodal semantic attributes for video captioning. 
An attention-based LSTM decoder has been proposed to pay attention to different semantic attributes from different modalities for enhancing sentence generation. 
In particular, our deep network has been optimized with semantic classification loss and video captioning loss simultaneously. 
Extensive experimental results have validated the effectiveness of our models 
on two standard video captioning datasets. 



\small
\bibliographystyle{IEEEbib}
\bibliography{icme2019template}
\end{document}